\documentclass[runningheads,a4paper]{llncs}

\usepackage{amssymb}
\usepackage{graphicx}

\usepackage{url}

\usepackage{wrapfig}
\usepackage{amsmath}
\usepackage{booktabs}
\usepackage{subcaption} 
\usepackage{color}
\usepackage{hyperref}
\usepackage{multirow}

\usepackage{dsfont} 

\DeclareMathOperator*{\argmax}{arg\,max} 

\newcommand{\Reals}{\mathds{R}}

\usepackage{diagbox} 

\begin{document}

\mainmatter  

\title{Transductive image segmentation:\\
Self-training and effect of uncertainty estimation}

\titlerunning{Transductive Image Segmentation}


\author{
Konstantinos Kamnitsas \inst{1} \and
Stefan Winzeck\inst{1} \and
Evgenios N. Kornaropoulos\inst{2,5} \and 
Daniel Whitehouse\inst{2} \and
Cameron Englman\inst{2} \and
Poe Phyu\inst{3} \and
Norman Pao\inst{4} \and \\
David K. Menon\inst{2} \and
Daniel Rueckert\inst{1,6} \and 
Tilak Das\inst{3} \and \\
Virginia F.J. Newcombe\inst{2} \and 
Ben Glocker\inst{1}}
\authorrunning{Kamnitsas et al.}

\institute{
Department of Computing, Imperial College London, UK \and
Division of Anaesthesia, Department of Medicine, University of Cambridge, UK \and
Dept Radiology, Cambridge Univ. Hospitals NHS Foundation Trust, UK \and
Dept Emergency Medicine, Cambridge Univ. Hospitals NHS Foundation Trust, UK \and
Clinical Sciences, Diagnostic Radiology, Lund University, Sweden \and
Klinikum rechts der Isar, Technical University of Munich, Germany
}

\toctitle{Lecture Notes in Computer Science}
\tocauthor{Authors' Instructions}
\maketitle



\begin{abstract}
    Semi-supervised learning (SSL) uses unlabeled data during training to learn better models. Previous studies on SSL for medical image segmentation focused mostly on improving model generalization to unseen data. In some applications, however, our primary interest is not generalization but to obtain optimal predictions on a specific unlabeled database that is fully available during model development. Examples include population studies for extracting imaging phenotypes. This work investigates an often overlooked aspect of SSL, transduction. It focuses on the quality of predictions made on the unlabeled data of interest when they are included for optimization during training, rather than improving generalization. We focus on the self-training framework and explore its potential for transduction. We analyze it through the lens of Information Gain and reveal that learning benefits from the use of calibrated or under-confident models. Our extensive experiments on a large MRI database for multi-class segmentation of traumatic brain lesions shows promising results when comparing transductive with inductive predictions. We believe this study will inspire further research on transductive learning, a well-suited paradigm for medical image analysis.
\end{abstract}


\section{Introduction}
\label{sec:intro}

The predominant paradigm for developing ML models is supervised training using labeled data. Because labels are scarce in many applications, these models often suffer from generalization issues on real-world heterogeneous data. To alleviate this, semi-supervised learning (SSL) \cite{chapelle2009semi}, aims to extract additional useful information from unlabeled data provided during training. Most commonly, the trained model is afterwards applied to \emph{new, previously unseen} unlabeled data at test time, predicts by \emph{induction}, and its generalization is measured by the accuracy on this new data. Model predictions made for the unlabeled data during training, as part of the optimization, are commonly discarded. For certain applications, however, the objective is to obtain predictions on a particular unlabeled database which may be \emph{fully available at time of model development}, rather than generalization to other data afterwards. Applications of this type include population analysis on retrospective data to extract image phenotypes (e.g. size or location of pathology). In such a setting, it is desirable to \emph{directly} optimize predictions for the specific unlabeled data, which is the concept of \emph{transduction}. In this context, this study takes a fresh look on SSL and evaluates the potential of transduction for image segmentation, which has been previously under-explored.

\emph{Transductive learning} is a form of SSL. Formally, SSL approaches assume two databases are available during model development: a \emph{labeled} database $D_L\!=\!\{X_L \cup Y_L\}\!=\!\{(x_{L,i},y_{L,i})\}_{i=1}^{N}$ and an \emph{unlabeled} database $D_U\!=\!\{X_U\}\!=\!\{x_{U,i}\}_{i=1}^{M}$, with $y_{L,i}\! \in\! \mathcal{Y}\! =\! \{1,...,C\}$ and $C$ the number of classes. Common SSL approaches construct model $f_{\theta}(x)\!=\!p(y|x;\theta)\! \in \! \Reals^{C}$ and learn parameters $\theta$ such that it will approximate distribution $q(y|x)$ that generates the labels. This is commonly done by jointly minimizing a supervised cost $J_s(X_L,Y_L,f_{\theta})$ (e.g. cross entropy) and an unsupervised cost, $J_u(X_U,\theta)$. For SSL focusing on \emph{induction}, the goal is to obtain optimized model $f_\theta^{\prime}$ which \emph{generalizes} to new, unseen and unlabeled \emph{target} (or \emph{test}) data $D_T$.
In contrast, in SSL focusing on \emph{transduction}, it is assumed that the unlabeled \emph{target} data is available during training (i.e. $D_U\!=\!D_T$). The goal is then to obtain the best possible predictions $Y_T^{\prime}=\{y^{\prime}_{T,i}\}_{i=1}^{M}$ that approximate the unknown true labels $Y_T\!=\!\{y_{T,i}\}_i^{M}$. Some methods, such as those employed in this study, can be used both for induction and transduction. However, in transduction the predictions on $D_T$ are directly optimized during training, and may thus be better than predictions via induction, as the latter is subject to degradation due to distributional shift and the generalization gap \cite{chapelle2009semi,vapnik1998statistical}.

Research on transductive SSL spans decades, with prime example the transductive SVM \cite{vapnik1998statistical}. In medical imaging, transduction has been mostly used for learning over graphs \cite{wang2016progressive,de2014tracing} and few-shot learning \cite{boudiaf2020transductive}. Related work also explored transductive learning of a meta-learner for combining predictions from multiple models \cite{zheng2019new}. We are instead interested in the common setting where a relation graph between samples does not exist, significant amounts of labels are available (not few-shot) and we investigate whether and how can transduction reliably benefit standard segmentation models.
In these settings, most prior work on SSL developed inductive methods and evaluated generalization to unseen data (e.g. \cite{bai2017semi,cui2019semi,yu2019uncertainty,kervadec2019curriculum,huang2018omni}).
Benefits by SSL were primarily shown when labeled data are very limited (few tens of images) and diminish with more labels \cite{bai2017semi,cui2019semi,yu2019uncertainty,kervadec2019curriculum}. In practical settings where more labeled data are available, the predominant paradigm is still supervised learning (e.g. large scale studies). Therefore, SSL methods that reliably offer improvements in such practical settings are still desirable.

This study makes the following contributions: (a) We present a theoretical analysis of the SSL framework of self-training via the lenses of Information Gain and reveal that learning benefits from well-calibrated or under-confident predictive uncertainty. This motivates us to introduce model ensembling within the framework, which is currently the most reliable method for obtaining calibrated uncertainty. (b) We assess quality of predictions obtained via transductive SSL and show consistent improvements over induction with our proposed framework. (c) We perform extensive evaluation on a large multi-center database for the challenging task of multi-class segmentation of traumatic brain injuries (TBI), including a blinded assessment via manual segmentation refinement, and demonstrate that transductive SSL with our framework can provide consistent improvements over induction, even when hundreds of labeled data are used. We believe this study will motivate further research into transductive learning, which is a suitable, but currently overlooked, paradigm for medical image analysis.

\section{Methodology}
\label{sec:main_methods}

We analyse and improve \emph{self-training} \cite{scudder1965probability}, a popular framework for inductive SSL with neural networks \cite{lee2013pseudo,bai2017semi}. Hence its transductive potential is of interest.

\subsection{Transductive learning via self-training}
\label{subsec:pseudolabels}

\begin{figure}[t]
\centering
\begin{subfigure}[b]{0.17\textwidth}
	\centering
	\includegraphics[clip=true, trim=0pt 0pt 0pt 0pt, width=1.0\textwidth]{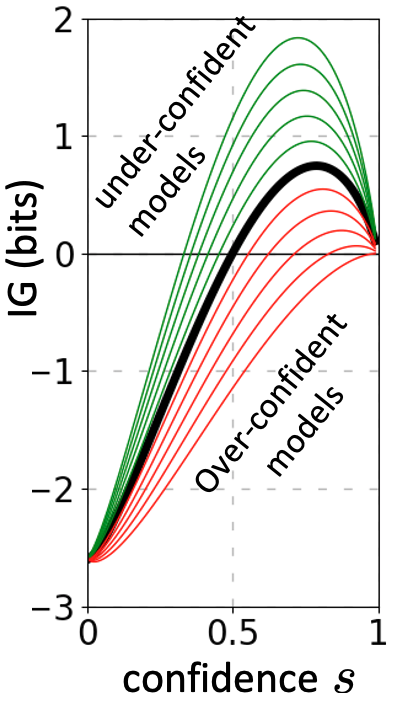}
	\caption{}
	\label{fig:info_gain}
\end{subfigure}
\begin{subfigure}[b]{0.81\textwidth}
	\centering
	\includegraphics[clip=true, trim=0pt 0pt 0pt 0pt, width=1.0\textwidth]{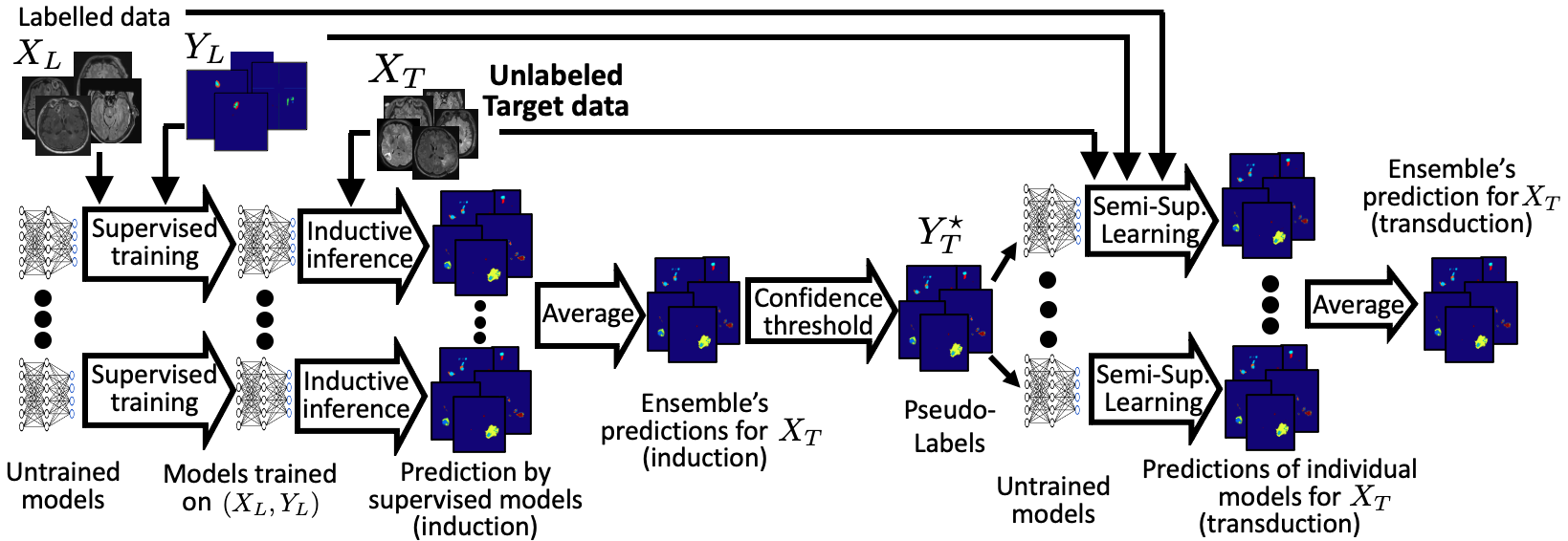}
	\caption{}
	\label{fig:pseudolabel_framework}
\end{subfigure}
\caption{a) Expected IG when a sample predicted with confidence $s$ is assigned a $1$-hot pseudo-label can be computed as $IG\!=\!\delta s H\! -\! (1\! -\! \delta s) H $, where $H$ the entropy in the predicted class-posteriors. Lines represent IG for different values of $\delta$. $\delta\!=\!1$ models perfect calibrated $s$, $0\!<\!\delta\!<\!1$ over-confident and $\delta\!>\!1$ under-confident. For illustration, here $H$ was calculated for $7$-class problem, with $s$ probability of most probable class and assuming $(1\!-\!s)/6$ for remaining classes.b) Self-training with ensembles for improved confidence calibration and learning. Predictions for target data are obtained via transduction in the end of 2nd training phase.}
\label{fig:methods}
\end{figure}

 In this framework, model $f_\theta$ (e.g. a neural net) is first trained via supervised cost $J_s(X_L,Y_L,\theta)$ to obtain $\theta^{\prime}$. Then, $f_{\theta^{\prime}}$ is applied to unlabeled data, which in our transductive setting are \emph{target} data $X_T$. The predicted labels $Y_T^{\prime}\!=\!\argmax_c(f_{\theta^{\prime}}(X_T))$ with highest class-posterior are then used as \emph{pseudo-labels} $Y_T^{\star}$ added to the training set. If the model provides confidence estimate $s_i$ for $i$-th sample, a \emph{confidence threshold} $t$ can be chosen, and only confident predictions ($s_i \! > \! t$) are used as pseudo-labels (in segmentation, uncertain pixels are masked out from the loss). The model is then re-trained on the extended training set minimizing $J_{total,ps} \! = \! J_s(X_L,Y_L,\theta) + \beta J_{s}(X_T,Y_T^{\star},\theta)$, with $\beta$ a hyper-parameter. This gives new parameters $\theta^{\prime\prime}$.
Inductive SSL strives to learn model $f_{\theta^{\prime\prime}}$ that will \emph{generalize} better than $f_{\theta^{\prime}}$ on \emph{new} data.
Instead, transductive learning focuses on predictions $Y_T^{\prime\prime}= f_{\theta^{\prime\prime}}(X_T)$ to improve them over $Y_T^{\prime}$ obtained via standard supervision.
We now analyse via Information Gain the framework, which reveals how uncertainty calibration of $f_{\theta^{\prime}}$ influences learning, guiding us to improve it.


\subsection{Analysing Information Gain and improving self-training}

Various works argue that SSL cannot gain information about the label generating process $p(y|x)$ solely from unlabeled data without priors or assumptions \cite{chapelle2009semi,scholkopf2013empirical,castro2020causality}. Then, \emph{how} does this framework benefit learning? A rigorous theory does not exist.
We conjecture that information useful for learning is given to the system via a \emph{subtle assumption} when pseudo-labels $Y_T^{\star}$ are created from class-posteriors.
We analyse this hypothesis through the lens of Information Gain (IG). IG is defined as the change of entropy $H$ of a system from a prior to a new state due to a new condition: $IG\!=\!H_{old}-H_{new}$.
Predicted posteriors for sample $x_i$ have entropy $H_i \! = \! -\!\sum_{c=1}^{C} p(c|x_i;\theta^{\prime}) \log p(c|x_i;\theta^{\prime})$. A 1-hot posterior has $0$ entropy. When a pseudo-label is created via $y_{T,i}^{\star}=\argmax_c p(c|x_i;\theta^{\prime})$, an \emph{assumption} is made that the class with highest posterior is correct and its posterior \emph{should} be 1. This generates $IG_i\!=\!H_i\!-\!0\!=\!H_i$ bits of info. If the prediction is correct, the pseudo-label matches the true unknown label, $y_{T,i}^{\star}=y_{T,i}$, and the obtained $H_i$ bits of information will facilitate learning. If the prediction is wrong, $y_{T,i}^{\star} \! \neq \! y_{T,i}$, the system is given $H_i$ bits of wrong information that hinder learning. We show next how confidence calibration relates to IG and how gains can be maximized.

A model's confidence score $s\! \in \! [0,1]$ is considered \emph{well calibrated} if $s \approx \frac{\sum_i \mathds{1}[y_i^{\prime}=y_i] \mathds{1}[s_i^{\prime}=s] }{\sum_i \mathds{1}[s_i^{\prime}=s]} \forall s $ (for discretized $s$). That is, confidence $s$ is equal to the ratio of correct ($y_i^{\prime}\!=\!y_i$) over all predictions made with confidence $s\!=\!s_i^{\prime}$.
As confidence score we use the model's maximum class posterior, $s_i \!=\! \max_c p(y\!=\!c|x_i;\theta)$ \cite{hendrycks2016baseline}. In this case, if we \emph{assume} model confidence is well-calibrated, we can estimate the \emph{expected} IG given to the system \emph{on average} when a sample predicted with confidence $s$ is assigned a 1-hot pseudo-label: $IG_{i,s}\! =\! s H_{i}\! -\! (1-s)\! H_i$. We plot this function in Fig.~\ref{fig:info_gain}. Standard neural nets, commonly used with self-training,
are over-confident in practice \cite{guo2017calibration,lakshminarayanan2016simple,hein2019relu}. Therefore we also plot the function $IG_{i,s}\! =\! \delta s H_{i}\! -\! (1-\delta s)\! H_i$, where $\delta$ models the level of over or under-confidence. Lower $\delta$ represents lower percentage of correct predictions than estimated by $s$ (over-confidence).
Fig.~\ref{fig:info_gain} shows that expected $IG_i$ is positive for samples above a certain confidence and negative below. This supports the common practice of using confidence threshold $t$ when creating pseudo-labels. Importantly, we find that for over-confident models, not only this threshold increases, but also the expected positive $IG_i$ decreases! This means that in self-training, using well-calibrated or under-confident models a) leads to higher gains and better learning than with over-confident models; b) allows a wider range of values for threshold $t$ without injecting negative IG to the system. The latter is practically important as choosing hyper-parameters is difficult for methods using unlabeled data \cite{oliver2018realistic}.

Guided by the above findings, we improve the framework by introducing the currently most reliable method for obtaining calibrated uncertainty with neural networks, model ensembling \cite{lakshminarayanan2016simple,jungo2020analyzing}.
Fig.~\ref{fig:pseudolabel_framework} shows the framework with network ensembles.
Individual models are trained with different weight initialization and data sampling, found sufficient for better uncertainty estimates (Fig.~\ref{fig:calibr}). Ensembling is also known to improve predictive accuracy, benefiting quality of pseudo-labels. We will demonstrate in Sec.~\ref{sec:experiments} empirical evidence aligned with our above two theoretical observations: Self-training via ensembling a) provides performance gains for different values of $t$ the magnitude of which follows trend aligned with what our theoretical analysis suggests (Fig.~\ref{fig:info_gain}), b) provides high performance for a wide range of $t$ making configuration easier and more reliable.

\section{Experimental evaluation}
\label{sec:experiments}

\subsection{Data and model configuration}
\label{subsec:data_model_config}

We evaluate transduction on multi-class TBI segmentation in the following setup.

\noindent\textbf{DB1:} This database was acquired in one clinic using 3 scanners. Consists of 180 subjects who underwent MRI including T1w, T2w, FLAIR, and SWI or GRE (SWI \& GE used interchangeably here). It includes $\sim\!85\%$ moderate/severe and $\sim\!15\%$ mild cases. DB1 labels are used for training in all following experiments.

\noindent\textbf{DB2:} This includes 101 subjects, acquired in 9 clinics with 4 scanner models, all different from DB1. T1w, T2w, FLAIR and SWI scans are available for all patients.
DB2 consists of $\sim\!35\%$ moderate/severe and $\sim\!65\%$ mild cases. These distribution shifts between DB1 and DB2 make for a challenging benchmark.
DB2 is used as \emph{target} data $D_T$ in following experiments, except for those in Sec.~\ref{subsec:manual_refine_eval}, where its labels are used for training and evaluation is done on DB3.

\noindent\textbf{DB3:} This database includes scans of 265 subjects from 14 sites (9 overlap with DB2, 5 new). Sequences and severity are similar to DB2. This database was unlabeled, kept unseen during model development. We used DB3 for blinded comparison of induction versus transduction after model development (Sec.~\ref{subsec:manual_refine_eval}).

\noindent\textbf{Manual annotations:} Experts annotated any abnormality visible in DB1 and DB2. Here, we consider 6 classes: core, oedema, petechial hemorrhages, intra-ventricular hemorrhages, non-TBI lesions, monitoring probe.

\noindent\textbf{Pre-processing:} We registered each image to the corresponding T1w, resampled to $1$mm isotropic resolution, and performed z-score intensity normalization.

\noindent\textbf{Main model:} We use a 3D Convolutional Network (CNN), DeepMedic, previously used for TBI segmentation \cite{kamnitsas2017efficient,monteiro2020multiclass}. We use the `wide' variant with default hyper-parameters (from: \url{https://github.com/deepmedic/deepmedic}, v0.8.4).

\noindent\textbf{Compared methods and configuration:} Besides self-training, we explore transduction with Entropy Minimization (EM), originally developed for inductive SSL \cite{grandvalet2005semi}. Moreover, training a CNN on pseudo-labels by ensemble with no confidence threshold $t$ (ST\_ens\_$0$ below) is equal to ensemble-distillation on unlabelled data \cite{bucilua2006model}.
Striving for reliability, we chose the studied methods for their simplicity, to avoid hyper-parameter tuning that is impractical in SSL \cite{oliver2018realistic}.
Except $t$ that we study below, the only hyper-parameter is the weight of unsupervised vs supervised cost, which we set to 1 for self-training and EM.

\subsection{Comparing supervised and transductive learning}
\label{subsec:results}

\begin{figure}[t]
	\centering
\begin{subfigure}[b]{0.32\textwidth}
	\centering
	\includegraphics[clip=true, trim=0pt 0pt 0pt 0pt, width=1.0\textwidth]{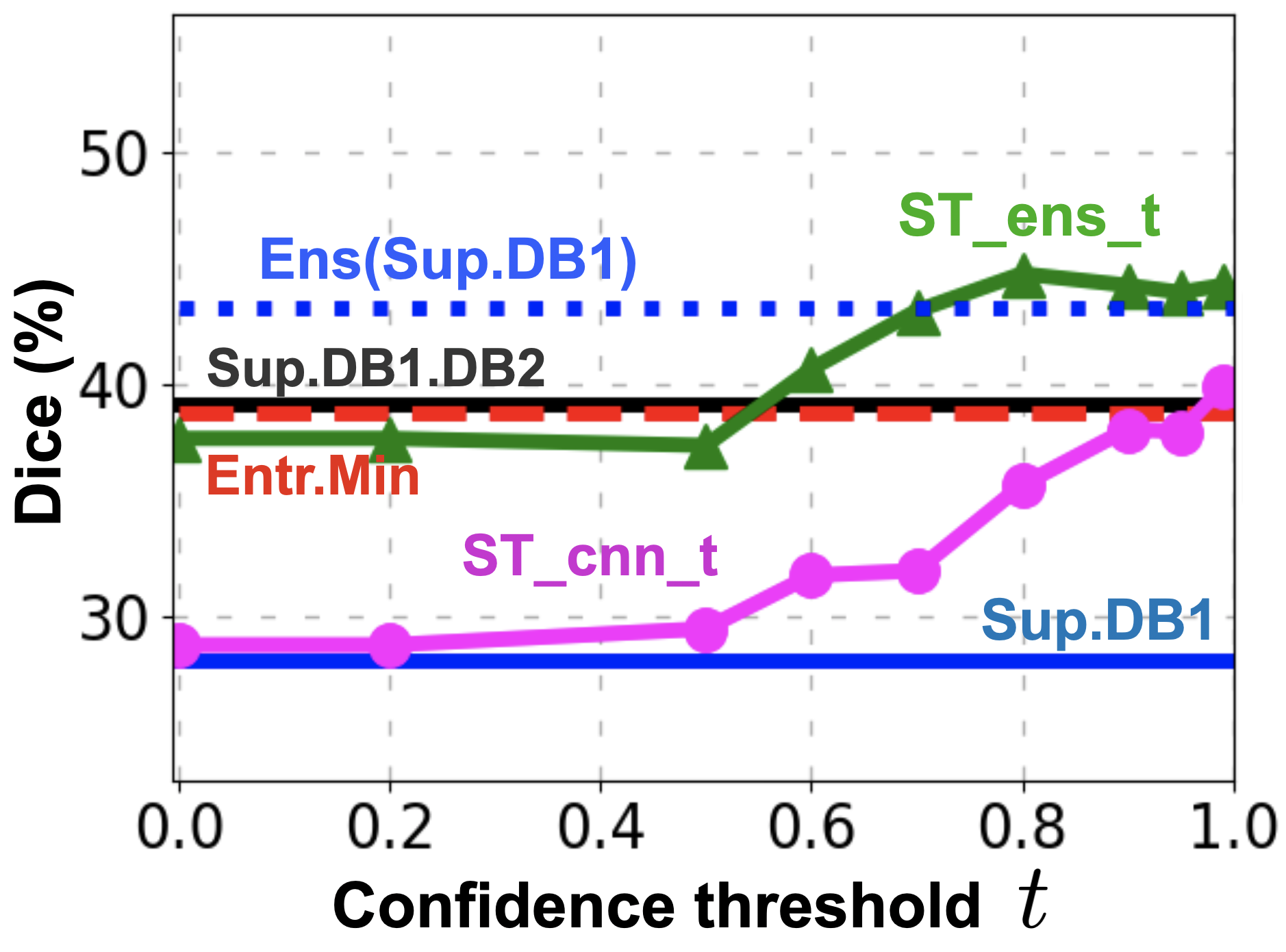}
	\caption{Single models}
	\label{fig:eval_singlemod}
\end{subfigure}
\begin{subfigure}[b]{0.32\textwidth}
	\centering
	\includegraphics[clip=true, trim=0pt 0pt 0pt 0pt, width=1.0\textwidth]{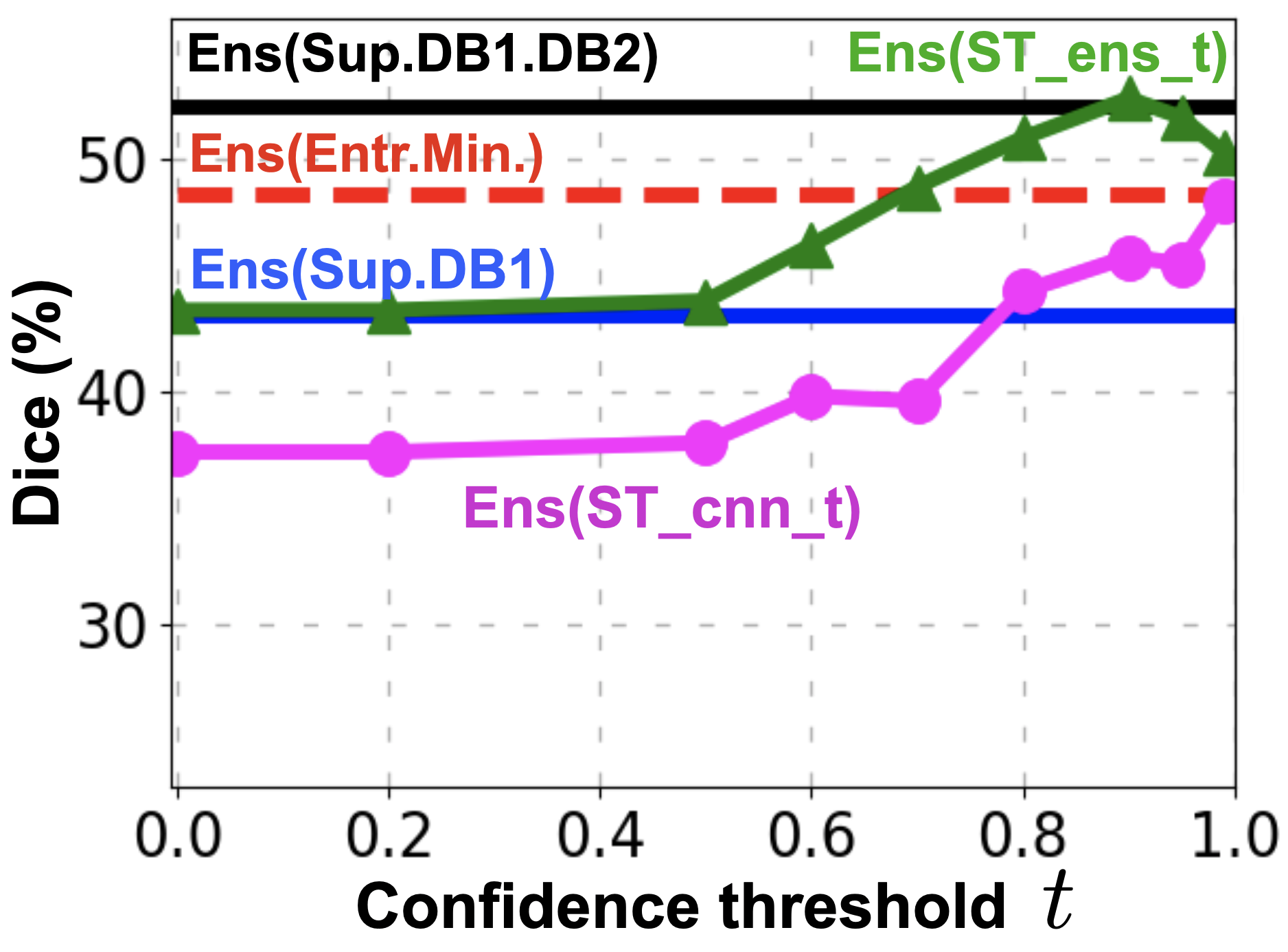}
	\caption{Ensembles}
	\label{fig:eval_ens}
\end{subfigure}
\begin{subfigure}[b]{0.28\textwidth}
	\centering
	\includegraphics[clip=true, trim=0pt 0pt 0pt 0pt,  width=1.0\textwidth]{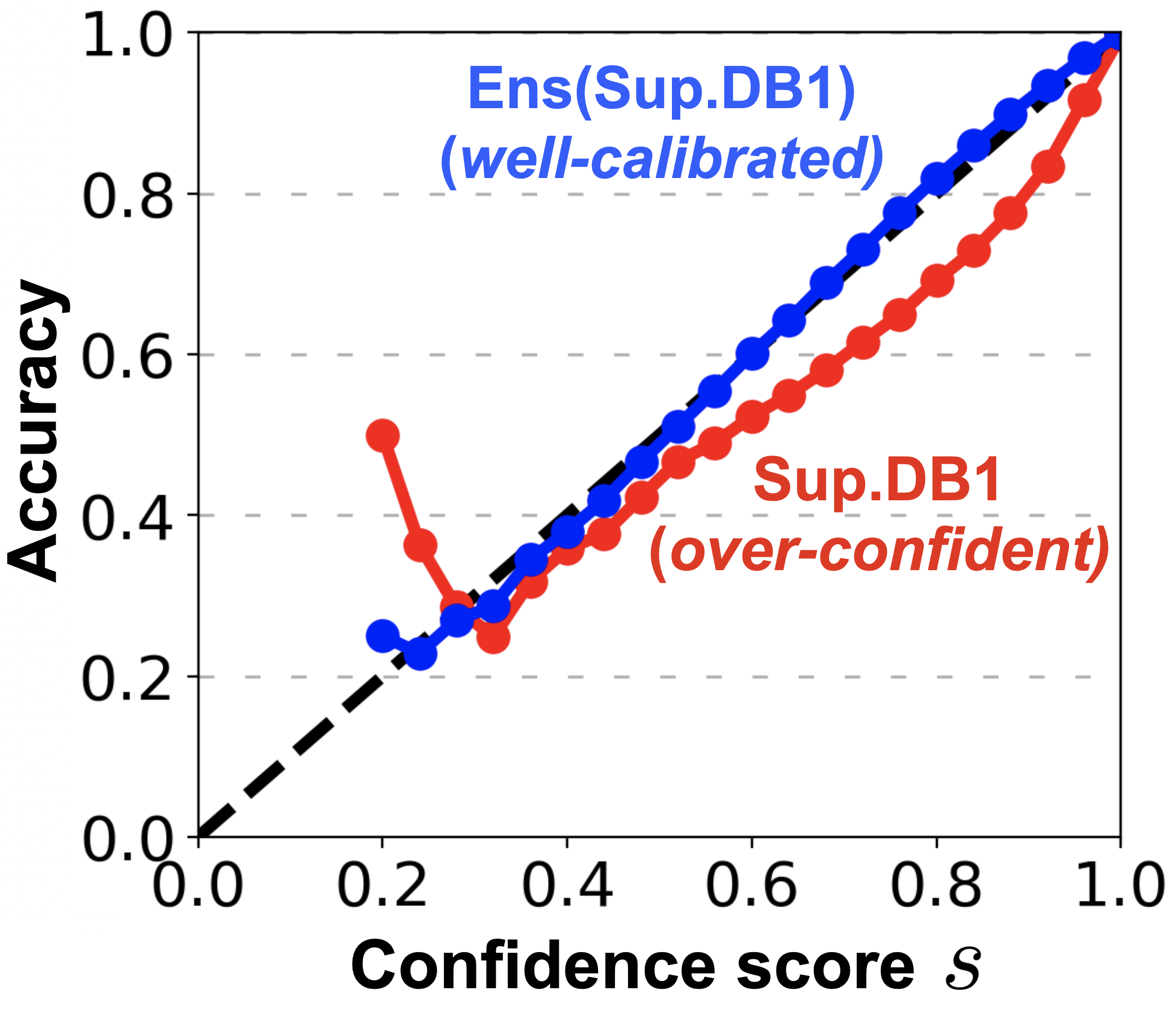}
	\caption{Calibration}
	\label{fig:calibr}
\end{subfigure}
\caption{(a,b) Average Dice (\%) over all 6 TBI classes achieved by different methods. Shown as function of confidence threshold $t$ for self-training methods. (c) Reliability diagram \cite{guo2017calibration} of a single CNN \emph{Sup.DB1} and an ensemble \emph{Ens(Sup.DB1)}.
}
\label{fig:evaluation}
\end{figure}

Each experiment below is repeated for $10$ seeds. Fig.~\ref{fig:evaluation} summarizes results. Shown is mean Dice over all classes and seeds, as space is limited for per-class analysis.

\noindent\textbf{Sup.DB1:} The supervised baseline are $10$ models trained with full supervision on DB1, evaluated on DB2. Models achieve $28.1\%$ Dice on average over all classes.
The low score is due to the tiny size of mild TBI lesions where few false positives greatly impact Dice (Fig.~\ref{fig:examples_results_tbi}), and domain shift between DB1-DB2.

\noindent\textbf{Sup.DB1.DB2:} We also evaluate how well a supervised model would perform if labels for DB2 were available for training. We train $10$ models using 100\%  DB1 and 80\% DB2 labels, and evaluate on remaining 20\% of DB2. We repeat for 5 folds of DB2. SSL without labels on DB2 is not expected to outperform this, but hopefully approach it.

\noindent\textbf{Entr.Min.:} We perform transductive learning by training $10$ models with supervision over DB1 and entropy minimization \cite{grandvalet2005semi} over unlabelled DB2 images. We evaluate predictions made on DB2 at end of training. Transduction with EM improves over baseline \emph{Sup.DB1} (Fig.~\ref{fig:eval_singlemod}).

\noindent\textbf{ST\_cnn\_$t$:} We evaluate transduction via self-training with pseudo-labels made by a single CNN. We create pseudo-labels using confidence threshold $t$ on posteriors of a supervised CNN (\emph{Sup.DB1}), and re-train a CNN on the extended training set. We repeat for $10$ models of \emph{Sup.DB1} and $7$ values of $t$ to investigate its effect. Using $t\!=\!0$ equals to using segmentations of \emph{Sup.DB1} as pseudo-labels because no prediction was less confident. Fig.~\ref{fig:evaluation} shows improvements over \emph{Sup.DB1}. Performance reaches EM only for best choice of $t$. High $t$ values are required, aligned with our theoretical findings about over-confident CNNs (Fig.~\ref{fig:info_gain}).

\noindent\textbf{Pseudo-labels from ensemble:} 
We assess performance of a CNN when trained with pseudo-labels derived from confidence estimates of an ensemble.
We make ensemble \textbf{Ens(Sup.DB1)} combining $10$ baseline CNNs (\emph{Sup.DB1}). Fig.~\ref{fig:calibr} shows the ensemble is indeed well-calibrated. We obtain pseudo-labels using threshold $t$ on ensemble's posteriors for DB2. We re-train $10$ CNNs using DB1 labels and DB2 pseudo-labels. Transductive predictions are obtained. We show average Dice of 10 seeds as \textbf{ST\_ens\_$t$}. We repeat for varying $t$.
CNN retrained using ensemble segmentations as pseudo-labels (\emph{ST\_ens\_$0$}, $t\!=\!0$, ala ensemble distillation \cite{bucilua2006model}) outperform \emph{ST\_cnn\_$t$}, thanks to the better segmentation of ensemble, which is not surprising. Most importantly, treating confidence via threshold $t$ offers performance gains that follow trend similar to what we derived for IG with well-calibrated models (Fig.~\ref{fig:info_gain}), supporting our theoretical results.
Close-to-optimal performance is achieved with wide range of $t$, in contrast to \emph{ST\_cnn\_$t$}, explained by the wider range of $t$ with positive IG of the former according to our findings (Fig.~\ref{fig:info_gain}). This facilitates the choice of a reliable $t$ value.
Finally, without labels for DB2, \emph{ST\_ens\_$t$} outperforms inductive \emph{Sup.DB1.DB2}, likely thanks to direct optimization of predictions on DB2 by transductive learning.

\noindent\textbf{Results by ensembles:} If ensemble is used to make pseudo-labels, then effective self-training should improve an ensemble. To assess this, we average transductive predictions from 10 models \emph{ST\_ens\_$t$} per $t$, obtaining transductive predictions of self-trained ensemble \textbf{Ens(ST\_ens\_$t$)} (Fig.~\ref{fig:pseudolabel_framework}). We compare with ensembles of CNNs re-trained with pseudo-labels from single CNNs (\textbf{Ens(ST\_cnn\_$t$)}), entropy minimization (\textbf{Ens(Entr.Min)}) and supervision on DB1$\cup$DB2 (\textbf{Ens(Sup.DB1.DB2)}).
Fig.~\ref{fig:eval_ens} shows that self-training improves the ensemble, with its transductive results (\emph{Ens(ST\_ens\_$t$)}) exhibit lower and less stable performance. Importantly, this holds for all values $t\!>\!0.5$, aligned with our findings for well-calibrated models (Fig.~\ref{fig:info_gain}). In contrast, ensembles re-trained on pseudo-labels from individual CNNs (\emph{Ens(ST\_cnn\_$t$)}) reach the baseline ensemble only for optimal $t$. Transductive ensemble trained via entropy minimization also surpasses the inductive ensemble. The best transductive ensemble approaches inductive \emph{Ens(Sup.DB1.DB2)} that requires labels for DB2.

\subsection{Comparing inductive and transductive semi-supervised learning}
\label{subsec:ind_vs_transd}

\begin{figure}[t]
	\centering
\begin{subfigure}[b]{0.50\textwidth}
	\centering
	\tiny
	\begin{tabular}{@{}lcc@{}}
	    \toprule
	      & \multicolumn{2}{c}{\textbf{Dice \%}} \\ 
		\textbf{Method}  & \textbf{Induct.} & \textbf{Transd.} \\
        \midrule 
		Sup.DB1            & $28.1$ & - \\
		Entr.Min.          & $36.4$ & $38.1$ \\
		ST\_ens\_0.7       & $39.7$ & $43.3$ \\
		ST\_ens\_0.9       & $41.9$ & $44.8$ \\
		\midrule
		Ens(Sup.DB1)       & $43.3$ & - \\
		Ens(Entr.Min.)     & $47.4$ & $48.4$ \\
		Ens(ST\_ens\_0.7)  & $48.8$ & $50.8$ \\
		Ens(ST\_ens\_0.9)  & $50.0$ & $53.3$ \\
		\bottomrule
	\end{tabular}
	\caption{Inductive vs transductive SSL}
	\label{tab:ind_vs_transd_TBI}
\end{subfigure}
\begin{subfigure}[b]{0.46\textwidth}
	\centering
	\includegraphics[clip=true, trim=0pt 0pt 0pt 0pt,  width=1.0\textwidth]{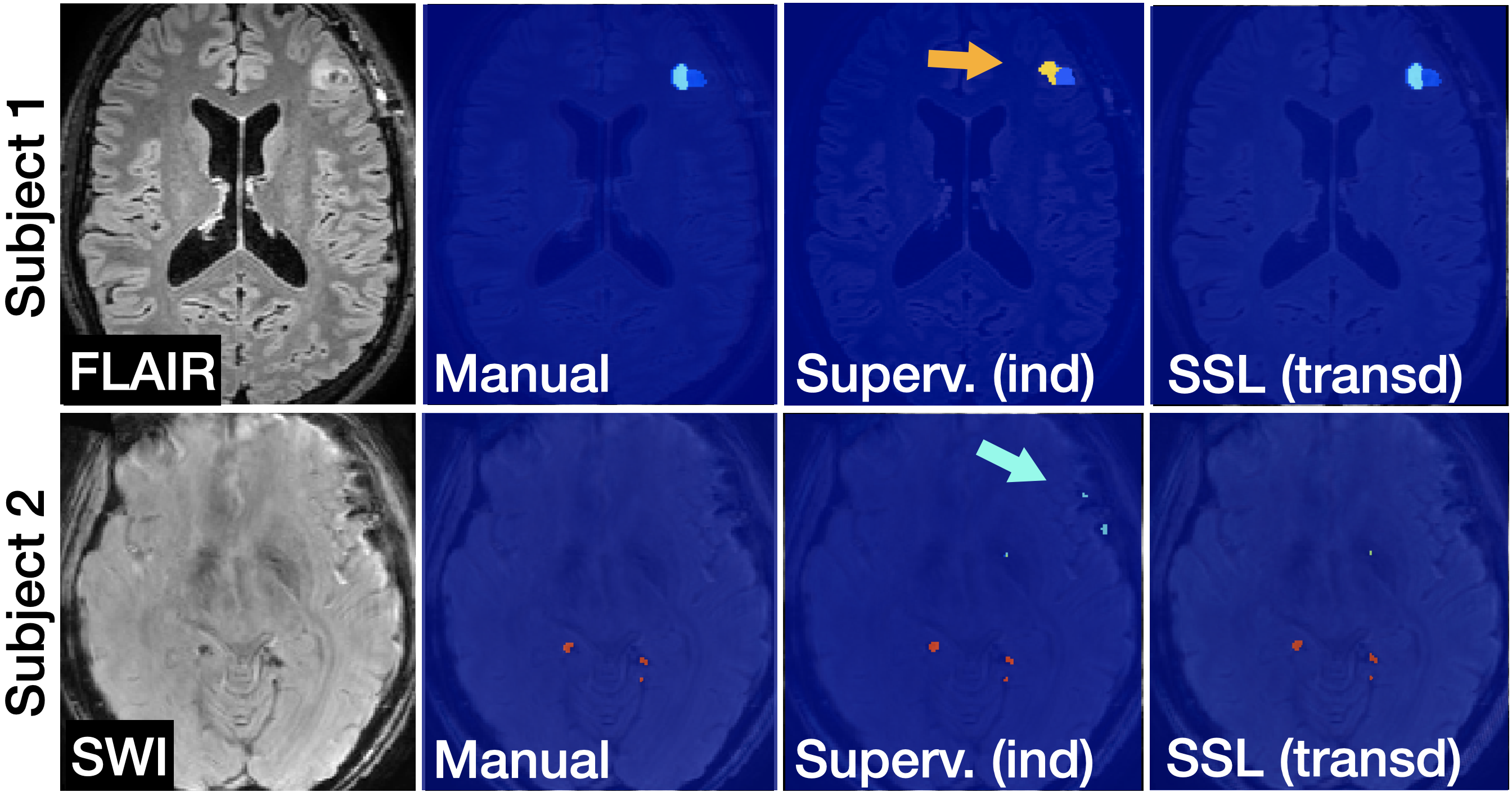}
	\caption{Example predictions}
	\label{fig:examples_results_tbi}
\end{subfigure}
\captionsetup{labelformat=empty}
\caption{Tab.3a) Comparison of transduction and induction. We show average Dice\% over all TBI classes and $10$ seeds. Differences (Ind/Transd) are significant ($p\!<\!0.05$). Fig.3b) Example predictions by induction (\emph{Ens(Sup.DB1)}) and transduction (\emph{Ens(ST\_ens\_0.7)}). Yellow arrow shows edema mistaken as related to monitoring probe. Cyan arrow shows false-positive TBI core. Both corrected by transductive SSL. Mild TBI lesions are often tiny, hence Dice degrades even with small errors.
}
\label{fig:ind_vs_transd_TBI_and_examples}
\end{figure}

We compare transduction and induction with the same model after self-training. To do this, we divide DB2 in 2 equal folds.
We start with pseudo-labels for DB2 from \emph{Ens(Sup.DB1)} (Sec.~\ref{subsec:results}). For 10 seeds, we perform self-training using 100\% DB1 labels and 50\% pseudo-labeled DB2.
After training, we obtain transductive predictions for the DB2 fold used in self-training and inductive predictions for the other fold.
We switch DB2 folds and repeat, to obtain these predictions for whole DB2 (\emph{ST\_ens\_t}).
Finally, we ensemble inductive and transductive predictions (separately) from 10 seeds, to assess ensemble performance (\emph{Ens(ST\_ens\_$t$)}). We repeat for $t\!=\!0.7$ and $0.9$ to assess along the wider effective range. We repeat similar procedure for the entropy minimization method. Results are shown in Table~\ref{tab:ind_vs_transd_TBI}. Transductive SSL outperforms inductive SSL in all cases.

\subsection{Blinded comparison via manual refinement of segmentations}
\label{subsec:manual_refine_eval}

\begin{table}[t]
	\begin{center}
	\caption{Dice\% between predicted and manually refined lesions on DB3. Transduction overlaps more than induction ($p\!<\!0.05$ for average over classes, \emph{Avg}).}
	\begin{tabular}{@{}lcccccc|c@{}}
	    \toprule
		\textbf{Method} & \textbf{Core} & \textbf{Oed.} & \textbf{Non-TBI} & \textbf{Probe} & \textbf{Petech.}  & \textbf{Intrav.} & \textbf{Avg.} \\
        \midrule 
		Induction, fully supervised: & $75.7$ & $77.9$ & $84.4$ & $85.7$ & $53.4$ & $83.9$ & $76.8$ \\
		Transduction, self-training: & $79.7$ & $79.8$ & $91.9$ & $82.4$ & $58.8$ & $83.9$ & $79.4$ \\
		\bottomrule
	\end{tabular}
	\label{tab:dice_manual_refine}
		\end{center}
\end{table}	

SSL is often found beneficial when labels are limited but its reliability is still questioned in practical settings, where supervised learning is preferred, e.g. large studies with more labels (Sec.~\ref{sec:intro}).
To address this, we compare transductive SSL with supervised training when using all 281 labeled cases of DB1$\cup$DB2, largest segmented cohort reported for MRI TBI, and predict 265 cases of multi-center DB3, which were unlabeled and held-out during model development for objective evaluation. We perform blinded comparison via manual segmentation refinement.

Because manual refinement is time consuming, we defined a process that requires only one refinement per image. We obtain \emph{inductive} predictions on DB3 using \emph{Ens(Sup.DB1.DB2)}, which is supervised on $DB1 \cup DB2$ (Sec.~\ref{subsec:results}). From these predictions we make pseudo-labels with $t=0.7$ (choice based on Fig.~\ref{fig:info_gain}). Using $DB1 \cup DB2$ labels and $DB3$ pseudo-labels we train an ensemble, similar to \emph{Ens(ST\_ens\_$0.7$)} (Sec.~\ref{subsec:results}), and obtain \emph{transductive} predictions. We then create an `in-between' segmentation per image by averaging inductive and transductive posteriors (ala ensembling). These latter segmentations were manually refined by clinicians with TBI expertise to meet the highest standards for follow-up studies on TBI phenotyping. Experts were unaware how the predictions were created.

Finally, we evaluated overlap of manually-refined segmentations with inductive and transductive predictions. Table~\ref{tab:dice_manual_refine} shows the results. Transduction clearly out-performs inductive supervised learning. These results show that transductive SSL can facilitate large-scale studies by reducing effort for manual refinement.

\section{Conclusion}

We explored the potential of transduction for segmentation. We showed with extensive experiments that if test data are available in advance, transduction outperforms induction by supervision or SSL. We also presented theoretical and empirical evidence that using well-calibrated or under-confident models facilitates self-training.
Future research should evaluate transduction on other data and perform analysis with multiple metrics, beyond Dice, to capture the method's effect more comprehensively. Having set the first stone with this study, future works should explore transductive segmentation with other SSL methods.
We believe these results will inspire further research in transductive SSL, which is well-suitable for retrospective medical image analysis studies.

\section{Acknowledgements}

This work received funding from the UKRI London Medical Imaging \& Artificial Intelligence Centre for Value Based Healthcare. VFJN is funded by an Academy of Medical Sciences / The Health Foundation Clinical Scientist Fellowship. DKM is supported by the National Institute for Health Research (NIHR, UK) through the Cambridge NIHR Biomedical Research Centre, and by a European Union Framework Program 7 grant (CENTER-TBI; Grant agreement 602150).

\bibliographystyle{splncs03}
\bibliography{references}

\end{document}